\documentclass[11pt,conference]{IEEEtran}
\IEEEoverridecommandlockouts
\usepackage{cite}
\usepackage{amsmath,amssymb,amsfonts}
\usepackage{algorithmic}
\usepackage{graphicx}
\usepackage{textcomp}
\usepackage{xcolor}
\usepackage{booktabs}

\def\BibTeX{{\rm B\kern-.05em{\sc i\kern-.025em b}\kern-.08em
    T\kern-.1667em\lower.7ex\hbox{E}\kern-.125emX}}
\begin{document}

\title{FoR-Net : Focus-on-Regions Network for Semantic Segmentation}
\author{
\IEEEauthorblockN{Sheng-Wei Chan}
\IEEEauthorblockA{
\textit{Dept. of Electrical Engineering} \\
\textit{Tamkang University} \\
412440330@o365.tku.edu.tw
}
\and
\IEEEauthorblockN{Hsin-Jui Pan}
\IEEEauthorblockA{
\textit{Dept. of Electrical Engineering} \\
\textit{Tamkang University} \\
412440314@o365.tku.edu.tw
}
\and
\IEEEauthorblockN{Jen-Shiun Chiang*}
\IEEEauthorblockA{
\textit{Dept. of Electrical Engineering} \\
\textit{Tamkang University} \\
jschiang@mail.tku.edu.tw
}
}
\maketitle

\begin{abstract}
This paper presents Focus-on-Regions Network (FoR-Net), an efficient semantic segmentation framework that explicitly focuses on hard regions through a selector-driven Top-K mechanism. Instead of relying on heavy global modeling, FoR-Net selectively enhances structurally informative regions. Multi-scale reasoning branches are introduced to aggregate spatial context efficiently. Experiments on the Cityscapes dataset demonstrate that FoR-Net achieves competitive performance while maintaining a lightweight architecture.
\end{abstract}


\section{Introduction}
Semantic segmentation is a fundamental task in computer vision, with applications in autonomous driving, robotics, and scene understanding. Early CNN-based methods achieve strong performance through local feature extraction and multi-scale context aggregation. More recently, Transformer-based approaches further improve global dependency modeling, but often require substantial computational resources. Despite these advances, challenging regions such as thin structures and object boundaries remain difficult due to over-smoothing and insufficient local detail preservation. These regions usually occupy only a small portion of the image.  However, they contribute disproportionately to segmentation errors, especially in computationally efficient architectures. Meanwhile, many existing segmentation frameworks still process all spatial locations uniformly, causing computational resources to be distributed across redundant background regions. This dense reasoning behavior becomes inefficient in high-resolution segmentation, where only a limited portion of pixels correspond to structurally challenging areas. Instead of uniformly processing all spatial regions, we propose FoR-Net, an efficient segmentation framework that allocates feature reasoning toward structurally informative regions through a selector-driven Top-K mechanism. Unlike conventional sparse attention methods that primarily reduce computational complexity through token sparsification, FoR-Net explicitly performs region-wise selection to emphasize challenging spatial regions. Experimental results on Cityscapes demonstrate strong segmentation performance under practical high-resolution training settings. The main contributions of this work are summarized as follows. First, we propose a selector-driven Top-K framework for semantic segmentation. Second, we introduce a multi-scale reasoning strategy that enhances informative regions without relying on computationally expensive global interaction. 

\section{Related Work}
Early semantic segmentation methods are mainly based on convolutional neural networks (CNNs). FCN~\cite{long2015fully} first introduced an end-to-end dense prediction framework, while DeepLab Series~\cite{chen2017deeplab, chen2017rethinking, chen2018encoder} and PSPNet~\cite{zhao2017pyramid} improve segmentation performance through atrous convolution and pyramid pooling. Later methods such as UPerNet~\cite{xiao2018unified}, HRNet~\cite{wang2020deep}, and OCRNet~\cite{yuan2020object} further enhance feature representation using multi-scale fusion, high-resolution features, and object-context modeling. Although these methods achieve strong accuracy, they often depend on complex context aggregation modules. Recently, Transformer-based segmentation methods such as SETR~\cite{zheng2021rethinking}, Swin Transformer~\cite{liu2021swin}, and SegFormer~\cite{xie2021segformer} have demonstrated strong global modeling capability. However, dense self-attention often introduces considerable computational overhead and redundant feature interaction in high-resolution segmentation. FoR-Net instead performs selective region-focused reasoning through a Top-K mechanism combined with multi-scale convolutions for efficient contextual aggregation.

\section{Proposed Method}
\subsection{Overview}

FoR-Net follows an efficient encoder-decoder architecture based on a ResNet-101 backbone with an output stride of 8. Given an input image $I \in \mathbb{R}^{H \times W \times 3}$, the backbone extracts hierarchical feature maps for region-focused reasoning and spatial refinement.

\begin{figure}[t]
\centering
\includegraphics[width=1.0\linewidth]{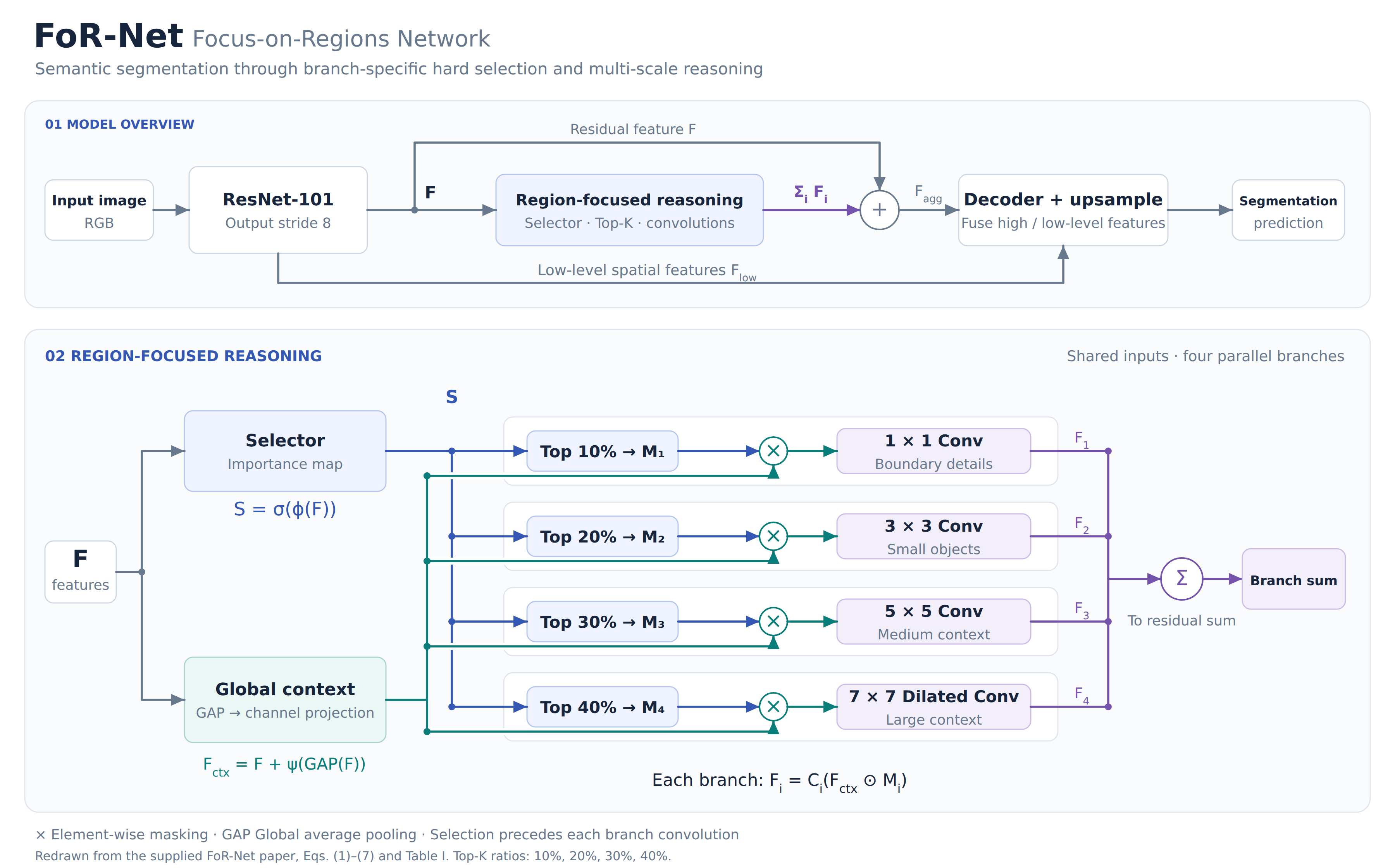}
\caption{Overview of FoR-Net. The model predicts an importance map and selects informative regions through a Top-K mechanism, followed by multi-scale reasoning and efficient decoding.}
\label{fig:fornet}
\end{figure}

As shown in Fig.~\ref{fig:fornet}, FoR-Net consists of three main components: (1) a selector module for predicting region-wise importance, (2) a Top-K mechanism for selecting informative regions, and (3) multi-scale reasoning branches for context aggregation.

\subsection{Region-Focused Reasoning}

To identify informative regions, a selector module predicts a spatial importance map from high-level features:

\begin{equation}
S = \sigma(\phi(F)),
\end{equation}
where $F \in \mathbb{R}^{H \times W \times C}$ denotes the high-level feature map extracted from the backbone, $H$, $W$, and $C$ represent the height, width, and channel dimension, respectively. $\phi(\cdot)$ denotes the selector network, and $\sigma(\cdot)$ is the sigmoid activation function. The predicted importance map $S \in \mathbb{R}^{H \times W}$ indicates the relative importance of each spatial location. Spatial locations associated with semantic ambiguity, structural complexity, or boundary inconsistency tend to produce higher selector responses during training.

A Top-K mechanism is applied to select informative spatial locations:

\begin{equation}
M = \text{TopK}(S),
\end{equation}
where $M \in \{0,1\}^{H \times W}$ denotes the binary selection mask generated from the predicted importance map. The Top-K operator retains the spatial locations with higher importance scores and suppresses the remaining locations.

The selected mask is used to enhance informative features:

\begin{equation}
F_{sel} = F \odot M,
\end{equation}
where $\odot$ denotes element-wise multiplication. This operation enhances selected spatial regions during feature propagation.

To incorporate global contextual information, an efficient global context branch is introduced:

\begin{equation}
F_{ctx} = F + \psi(\text{GAP}(F)),
\end{equation}
where GAP denotes global average pooling, and $\psi(\cdot)$ denotes an efficient channel transformation function implemented by a small projection layer.  It maps the global descriptor back to the feature channel dimension before being fused with the original feature map. The transformed global descriptor is fused with the original feature map to provide additional global context.

\subsection{Multi-scale Reasoning}

To capture multi-scale contextual information, FoR-Net employs parallel convolution branches with different receptive fields:

\begin{equation}
F_i = \mathcal{C}_i(F_{ctx} \odot M_i),
\end{equation}

\noindent where $M_i$ denotes the branch-specific Top-K mask for the $i$-th branch, and $\mathcal{C}_i(\cdot)$ denotes convolution operations with different receptive fields. Different branches are designed to capture contextual information at different spatial scales, as summarized in Table~\ref{tab:multiscale_roles}.

\begin{table}[t]
\caption{Semantic roles and Top-K selection strategies of different receptive fields in FoR-Net.}
\label{tab:multiscale_roles}
\centering
\scriptsize
\setlength{\tabcolsep}{2.5pt}
\begin{tabular}{ccc}
\toprule
\textbf{Kernel} & \textbf{Primary Role} & \textbf{Top-K} \\
\midrule
$1\times1$ & Boundary details & 10\% \\
$3\times3$ & Small objects & 20\% \\
$5\times5$ & Medium context & 30\% \\
$7\times7$ & Large context & 40\% \\
\bottomrule
\end{tabular}
\end{table}

The Top-K ratios are empirically selected according to validation performance and branch-specific receptive field characteristics. Smaller receptive fields focus on local structural details, while larger receptive fields provide broader contextual information.

The outputs are aggregated as:

\begin{equation}
F_{agg} = F + \sum_i F_i.
\end{equation}
The aggregated feature $F_{agg}$ combines information from multiple receptive-field branches while preserving the original high-level feature representation through residual connection.

Finally, the refined features are fused with low-level representations through an efficient decoder:

\begin{equation}
Y = \mathcal{D}(F_{agg}, F_{low}),
\end{equation}
where $F_{low}$ denotes low-level spatial features extracted from earlier backbone stages, and $\mathcal{D}(\cdot)$ denotes the efficient decoder. The decoder combines high-level semantic features and low-level spatial details to generate the final segmentation prediction.

The overall training objective is defined as:

\begin{equation}
\mathcal{L} = \mathcal{L}_{CE} + \lambda_1 \mathcal{L}_{Dice} + \lambda_2 \mathcal{L}_{sel}.
\end{equation}
where $\mathcal{L}_{CE}$ denotes the cross-entropy loss for semantic classification, $\mathcal{L}_{Dice}$ denotes the Dice loss for improving region-level consistency, and $\mathcal{L}_{sel}$ encourages the selector to emphasize challenging spatial regions, particularly around object boundaries and fine-grained structures. The selector is jointly optimized with the segmentation objective and implicitly learns to allocate higher responses to spatial regions that contribute more significantly to segmentation errors. $\lambda_1$ and $\lambda_2$ are balancing coefficients for the Dice and selector losses.

\section{Experiments}

\subsection{Experimental Setup}

We evaluate FoR-Net on the Cityscapes dataset~\cite{cordts2016cityscapes}. All experiments are conducted under practical single-GPU training settings with approximately 16GB GPU memory usage. The model adopts a ResNet-101 backbone with an output stride of 8 and is trained using AdamW with an initial learning rate of $1\times10^{-4}$. Following common practice, random scaling, cropping, and horizontal flipping are applied for data augmentation. Under a high-resolution training configuration with a crop size of $768\times768$ and output stride 8, FoR-Net supports a batch size of 8 on a single GPU. 

\subsection{Quantitative Results}
Table~\ref{tab:main_results} reports the validation mIoU on Cityscapes. Compared with representative segmentation architectures, FoR-Net achieves a favorable efficiency-performance trade-off by selectively enhancing informative regions instead of uniformly processing all spatial locations. Reported results are taken from the original papers.

\begin{table}[t]
\caption{Comparison with representative semantic segmentation methods on the Cityscapes validation set.}
\label{tab:main_results}
\centering
\scriptsize
\setlength{\tabcolsep}{2.2pt}
\begin{tabular}{lcccc}
\toprule
\textbf{Method} & \textbf{Backbone} & \textbf{Param. (M)} & \textbf{GFLOPs} & \textbf{mIoU} \\
\midrule
DeepLabV3~\cite{chen2017rethinking} & ResNet-101 & 60.22 & 254.0 & 77.23 \\
PSPNet~\cite{zhao2017pyramid} & ResNet-101 & 65.60 & 256.0 & 78.40 \\
DANet~\cite{fu2019dual} & ResNet-101 & 66.47 & 289.0 & 78.8 \\
PSANet~\cite{zhao2018psanet} & ResNet-101 & 78.13 & 264.0 & 78.6 \\
DeepLabV3+~\cite{chen2018encoder} & ResNet-101 & 84.74 & 348.0 & 79.55 \\
ANNNet~\cite{zhu2019asymmetric} & ResNet-101 & 62.86 & 348.0 & 80.5 \\
DNL~\cite{yin2020disentangled} & ResNet-101 & 71.49 & 343.0 & 80.5 \\
CCNet~\cite{huang2019ccnet} & ResNet-101 & 66.13 & 276.0 & 80.5 \\
DGMNet~\cite{chan2026breaking} & ResNet-101 & 53.00 & 224.1 & 80.8 \\
\midrule
\textbf{FoR-Net} & \textbf{ResNet-101} & \textbf{55.72} & \textbf{243.6} & \textbf{80.5} \\
\bottomrule
\end{tabular}
\end{table}

\begin{figure}[t]
\centering

\includegraphics[width=0.36\linewidth]{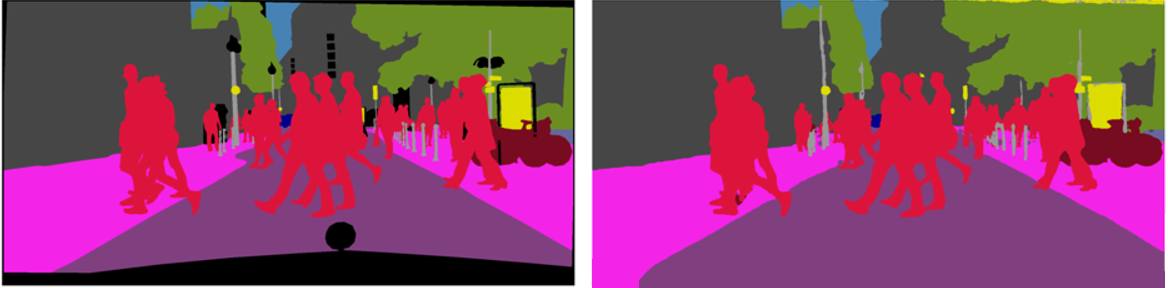}
\hspace{0.04\linewidth}
\includegraphics[width=0.36\linewidth]{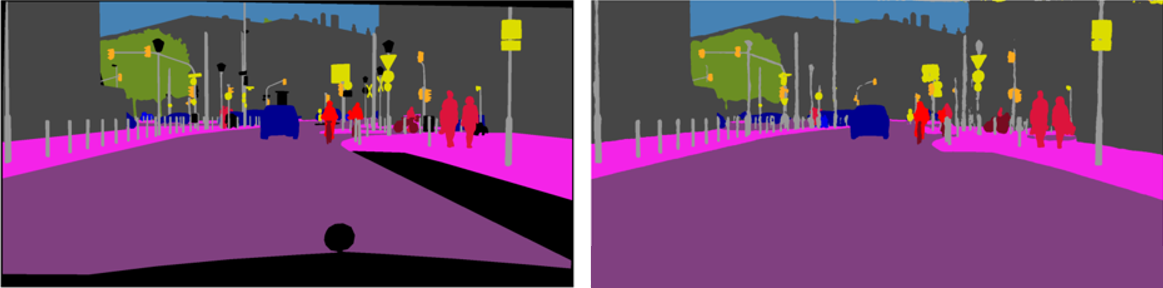}

\vspace{3pt}

\includegraphics[width=0.36\linewidth]{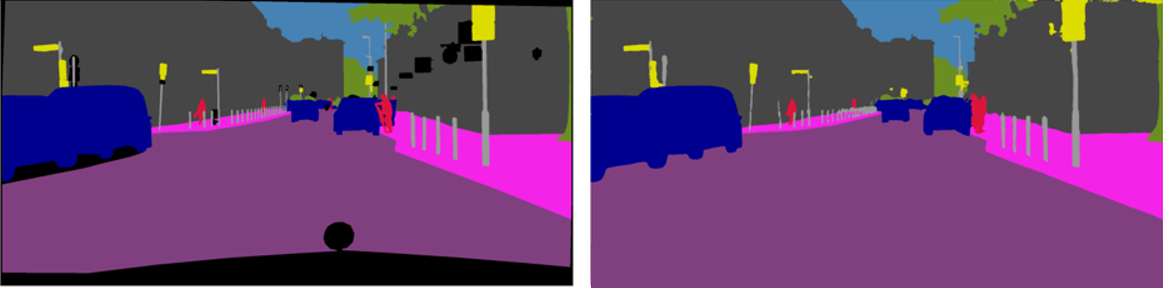}
\hspace{0.04\linewidth}
\includegraphics[width=0.36\linewidth]{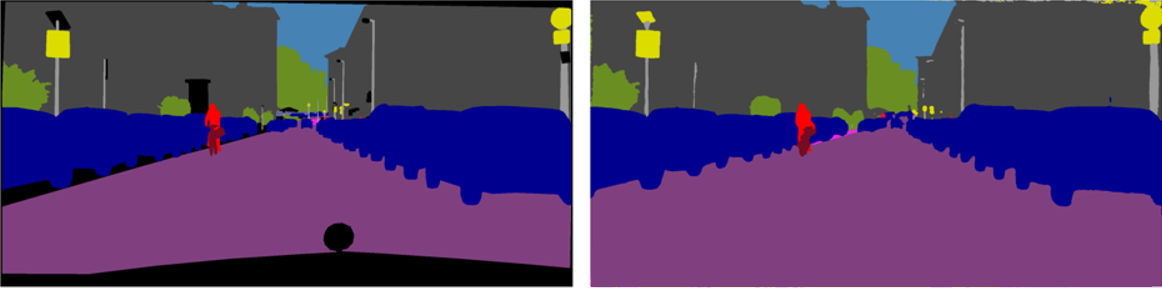}

\caption{Qualitative comparison on the Cityscapes validation set. Each group shows the input image, ground truth, ResNet-101 baseline, and FoR-Net prediction from left to right. FoR-Net produces more consistent predictions in challenging regions.}
\label{fig:qualitative}
\end{figure}

FoR-Net improves predictions in challenging regions. As illustrated in Fig.~\ref{fig:qualitative}, FoR-Net preserves thin structures and object boundaries more effectively than baseline predictions, reducing over-smoothing artifacts in challenging regions.

\subsection{Ablation Study}

We further investigate the impact of receptive fields by varying the dilation rate in the $7\times7$ convolution branch. As shown in Table~\ref{tab:ablation}, larger dilation rates consistently improve segmentation performance, indicating that expanded receptive fields help capture more effective multi-scale contextual information.

\begin{table}[t]
\caption{Ablation study on dilation rates.}
\label{tab:ablation}
\centering
\small
\begin{tabular}{cc}
\toprule
Dilation Rate & mIoU (\%) \\
\midrule
1 & 77.42 \\
2 & 78.59 \\
4 & 79.44 \\
8 & 80.16 \\
16 & 80.49 \\
\bottomrule
\end{tabular}
\end{table}

We additionally observe that excessively sparse activation in large receptive field branches may negatively affect semantic context modeling. In particular, assigning overly small Top-K ratios to the $7\times7$ branch restricts contextual propagation and leads to degraded semantic completeness in ambiguous regions. Empirically, larger receptive field branches benefit from broader spatial coverage, while smaller receptive fields are more suitable for highly selective boundary-sensitive reasoning. 

\section{Conclusion}
In this paper, we presented FoR-Net, an efficient semantic segmentation framework based on a selector-driven Top-K mechanism for region-focused reasoning. By selectively focusing computation on important regions, FoR-Net achieves strong segmentation performance with efficient computation. Experiments on the Cityscapes dataset show improved consistency in challenging regions, especially for fine-grained structures under resource-constrained settings. Future work will explore adaptive region selection and evaluation on more diverse benchmarks.

\section*{Acknowledgment}
\noindent This research work is partially supported by the National Science and Technology Council, Taiwan, under grant number 114-2221-E-032-011.

\bibliographystyle{IEEEtran}
\bibliography{refs}

\end{document}